# SENSE: SELF-SUPERVISED NEURAL EMBEDDINGS FOR SPATIAL ENSEMBLES

**HAMID GADIROV*, LENNARD MANUEL, STEFFEN FREY**

*University of Groningen,*

*Netherlands*

** h.gadirov@rug.nl*

**ARTICLE INFO**

**ABSTRACT**

*Analyzing and visualizing scientific ensemble datasets characterized by high dimensionality and structural complexity remains a significant challenge. While dimensionality reduction methods and autoencoders are widely used for feature extraction, their performance often degrades in high-dimensional settings. In this work, we propose an enhanced autoencoder framework that integrates clustering and contrastive loss functions into the latent space to improve the interpretability and visualization of scientific ensemble data. Clustering is guided by a soft silhouette score, encouraging compact and well-separated latent representations. To address the presence of unlabeled data, EfficientNetV2 is employed to generate pseudo-labels for partially unlabeled ensembles. The model is trained by jointly optimizing reconstruction, clustering, and contrastive objectives, resulting in improved grouping of similar samples and clearer separation of distinct structures in the latent space. UMAP is subsequently applied to the learned embeddings to produce two-dimensional visualizations, which are quantitatively evaluated using silhouette scores. We evaluate multiple autoencoder variants on two scientific ensemble datasets: subsurface channel structures generated via Markov chain Monte Carlo simulations and droplet-impact dynamics on a liquid film. The results show that incorporating clustering or contrastive objectives yields marginal but consistent improvements over baseline autoencoders.*

## 1. Introduction

In recent years, the growth of high-dimensional scientific ensemble datasets has presented both opportunities and challenges for data analysis and visualization [1]. Scientific ensembles, characterized by their complex and multi-dimensional nature, contain valuable insights that can assist in decision-making processes across various domains, from climate modeling to healthcare diagnostics [2]. However, extracting meaningful features from these datasets remains a difficult task due to their complexity.

To make these complex ensemble datasets more understandable, dimensionality reduction techniques can be applied. However, these techniques struggle to uncover the structures in high-dimensional datasets. Thus, we first apply feature extraction through the use of (variational) autoencoders. These models extract the most relevant features from the datasets, after which dimensionality reduction techniques can be used to obtain a more intuitive visualization.

Combining these two methods has shown promising results, but we hope to achieve better clustering within this visualization [3].

In this paper, we propose a novel approach for clustering scientific ensemble datasets by combining the strengths of autoencoder-based feature extraction with a dedicated clustering and contrastive loss function. Our method aims to extract relevant features of scientific data while simultaneously encouraging the forming of distinct clusters in the latent space. By jointly optimizing the reconstruction objective during training as well as the cluster separability enforced by either the clustering or contrastive loss, our approach offers a framework to obtain a better understandable visualization.

We implement a soft silhouette score, which is a differentiable version of the silhouette score. This score is implemented as a clustering loss alongside the reconstruction loss of a (variational) autoencoder during training, ensuring that the data will form more compact clusters while also preserving the original data's information. This clustering loss will also be compared to a contrastive loss function. Contrastive loss aims to bring instances of the same class closer together while pushing apart the instances of different classes, ensuring that similar data will be grouped in the latent space. This clustering during training can be performed on mostly unlabelled datasets because EffNetV2 will be used to first train the model on the manually labelled part of the ensemble datasets, after which pseudo-labels for the unlabelled part of the datasets will be generated, making this a semi-supervised problem.

Once the training is finished, we will perform dimensionality reduction to further reduce the latent space to a 2D visualization by performing dimensionality reduction: specifically by utilizing UMAP. The resulting visualizations will be evaluated by their silhouette score and compared to similar models, with and without a clustering or contrastive loss. In our experiments, we used two ensemble datasets: Markov Chain Monte Carlo and Drop Dynamics [4], [5].

In Section 2, we give a brief overview of related works. Afterwards, we describe the methodology used for this paper in Section 3. Then, we move on to the results and discussion in Section 4. We conclude our findings in Section 5. Finally, we discuss future works in Section 6.

This work is based on the master's internship project by Lennard Manuel titled "Autoencoder-based semi-supervised dimensionality reduction and clustering for scientific ensembles" from the University of Groningen [40].

## 2. Related Work

In this section, we briefly describe what research has been done previously in the fields of autoencoder-based feature extraction, deep clustering, and contrastive learning.

**Autoencoder-based Feature extraction.** Autoencoders have become instrumental in the field of feature extraction due to their ability to learn efficient, compressed representations of high-dimensional data. Ardelean et al. propose autoencoders as a feature extraction method for spike sorting, the process of grouping spikes of distinct neurons into their respective clusters [6]. Autoencoders are also widely used in computer vision. Nayak et al. use a deep autoencoder to help detect brain tumors in medical images [7]. Chen et al. propose a convolutional autoencoder to help in detecting and analyzing long nodules [8]. Solomon et al. use autoencoders to develop a face verification system [9]. Furthermore, Variational Autoencoders are also useful in this process. Tian et al. developed the Pyramid-VAE-GAN network to assist in image inpainting [10].

More recently, Yin et al. introduced ENTIRE, an autoencoder-based framework that extracts structure-aware feature representations from time-dependent volumetric data and combines them with rendering parameters to accurately predict volume rendering time, enabling dynamic parameter adaptation and load balancing in visualization pipelines [37].

**Deep clustering.** Deep clustering refers to the process of integrating deep learning networks with clustering meth- ods. It helps in transforming the input data such that clusters will try to form within the latent space [11]. In the paper "Deep clustering using the soft silhouette score: towards compact and well-separated clusters" Vardakas et al. introduce a probabilistic formulation of the silhouette score to complement their autoencoders' reconstruction loss with a clustering loss [12]. They use a Radial Basis Function model as a clustering network to predict the probabilities with which they calculate the soft silhouette score. They show promising results on the EMNIST datasets. Xie et al. propose the Deep Embedding Clustering (DEC) method, which also optimizes both the reconstruction and clustering objective using deep neural networks [13]. They use the KL divergence as their clustering loss. Guo et al. adapt the DEC method and develop the Improved Deep Embedding Clustering (IDEC) method [14]. This method simultaneously optimizes the reconstruction and clustering objective during the training phase, whereas DEC pre-trains on the reconstruction objective, after which it optimizes the clustering objective. Yang et al. also propose their own method: the Deep Clustering Network (DCN) [15]. This method tries to optimize the clustering objective using k -means on the embedded space.

**Contrastive loss.** Contrastive learning is a deep learning technique that is effective in creating separation between different classes. Zhou et al. for example propose a contrastive autoencoder (CAE-AD) for anomaly detection in multivariate time series [16]. Luo et al. also combine contrastive learning with an autoencoder to assist in out-of-distribution detection [17]. Lopez-Avila et al. combine a denoising autoencoder with contrastive learning to help fine-tune their transformer models [18]. Contrastive learning also has its place in the medical world: Cao et al. propose ContrastNet, which combines prototypical contrastive learning with an autoencoder to create an unsupervised feature learning network for hyperspectral classification [19].

**Ensemble data analysis.** Modern scientific simulations and measurements often generate large spatio-temporal ensembles, where each member represents a different realization of the same phenomenon under varying parameters or initial conditions. Extracting meaningful low-dimensional structure from such ensembles is crucial for tasks such as uncertainty quantification, pattern discovery, and interactive visualization. Autoencoder-based methods are particularly well suited for this setting because they can learn compact latent representations that preserve salient spatial and temporal structures while remaining flexible across different simulation domains. In the context of ensemble visualization, autoencoder architectures have been systematically evaluated as feature extractors and dimensionality reduction models for spatial ensembles, with a focus on how architectural choices affect projection quality and the expressiveness of the resulting embeddings [35]. These studies demonstrate that appropriate encoder–decoder configurations can yield latent spaces in which ensemble members cluster according to physically meaningful behaviors, thereby supporting downstream visual analysis and clustering.

Beyond static dimensionality reduction, recent work has explored deep learning for learning flow fields and reconstructing missing temporal information in ensemble data. FLINT introduces a learning-based approach for flow estimation and temporal interpolation that reconstructs velocity fields and scalar quantities in 2D+time and 3D+time ensembles, enabling high-quality in-between time steps without strong domain assumptions [34]. HyperFLINT extends this idea with a hypernetwork that conditions on simulation parameters, improving generalization across different ensemble configurations and supporting parameter-aware interpolation quality [33]. Together, these methods illustrate how neural networks can capture both spatial and temporal coherence in ensembles and use this knowledge to fill temporal gaps, generate dense time sampling, and support fluid, artifact-free animation and analysis [38].

These developments are part of a broader trend toward machine-learning-driven ensemble data analysis in scientific visualization. A recent comprehensive study on machine learning for scientific visualization discusses autoencoder-based dimensionality reduction, learning-based flow estimation, and hypernetwork-based adaptation as complementary building blocks for analyzing complex spatio-temporal ensembles [36]. In parallel, autoencoders and related deep architectures have been applied to other ensemble-related tasks, such as learning low-dimensional probabilistic representations of ensemble forecast fields and leveraging variational autoencoders for ensemble visualization and uncertainty exploration in meteorology. The usefulness of such learned latent representations is further highlighted by applications beyond "pure" data analysis. For example, ENTIRE uses deep features derived from volumetric data and camera parameters to predict volume rendering time, enabling dynamic parameter adaptation and more stable interactive performance for time-dependent volume data [37]. Collectively, these works underline that autoencoder-based representations and related deep learning techniques provide a powerful and versatile foundation for ensemble data analysis, supporting not only visualization and clustering but also performance prediction and system-level steering in complex simulation workflows.

**Recent advances in Artificial Intelligence.** Recent advances in large language models (LLMs) and foundation models have begun to influence scientific data analysis workflows, including ensemble-based visualization and simulation analysis. While LLMs are primarily developed for language tasks, their underlying architectural and system-level innovations—such as attention mechanisms, scalable training paradigms, and modular model composition—are increasingly relevant for handling large, heterogeneous scientific datasets. In particular, LLM-driven interfaces and multimodal models show promise for assisting exploratory analysis, metadata reasoning, and interactive steering of ensemble simulations by coupling learned representations with semantic context. As scientific datasets continue to grow in size and complexity, real-time optimization and scalability have become central concerns. Techniques such as mixed-precision training, low-rank adaptation (LoRA), and model pruning significantly reduce computational overhead while preserving model performance [39]. System-level frameworks including vLLM and DeepSpeed further enable efficient training and inference at scale, while sparse expert architectures such as Switch Transformers demonstrate how conditional computation can reduce resource usage without sacrificing expressiveness. Complementary approaches based on reinforcement learning allow models to adapt their complexity dynamically in response to available computational budgets. Moreover, distributed strategies such as federated learning and ZeRO improve scalability and collaboration across institutions handling large-scale ensemble datasets. Finally, emerging work on agentic AI systems—in which autonomous agents coordinate learning, inference, and analysis tasks—opens new directions for ensemble data processing. When combined with containerization technologies such as Docker, these agents can be deployed as modular, reproducible components that manage data ingestion [41, 42], model execution, and visualization pipelines in a flexible and scalable manner, supporting robust, system-level integration of learning-based ensemble analysis.

## 3. Methodology

In this section, we discuss and explain all steps of the pipeline of our approach, which is shown in Figure 1. First, the ensemble data is preprocessed, after which we generate the pseudo- labels using EffNetV2. Following this, feature extraction is performed with the (variational) autoencoders using reconstruction loss combined with either clustering or contrastive loss. Then, the latent space is projected to a 2D space using UMAP, which is then evaluated by their silhouette scores. Then, all the results are compared.

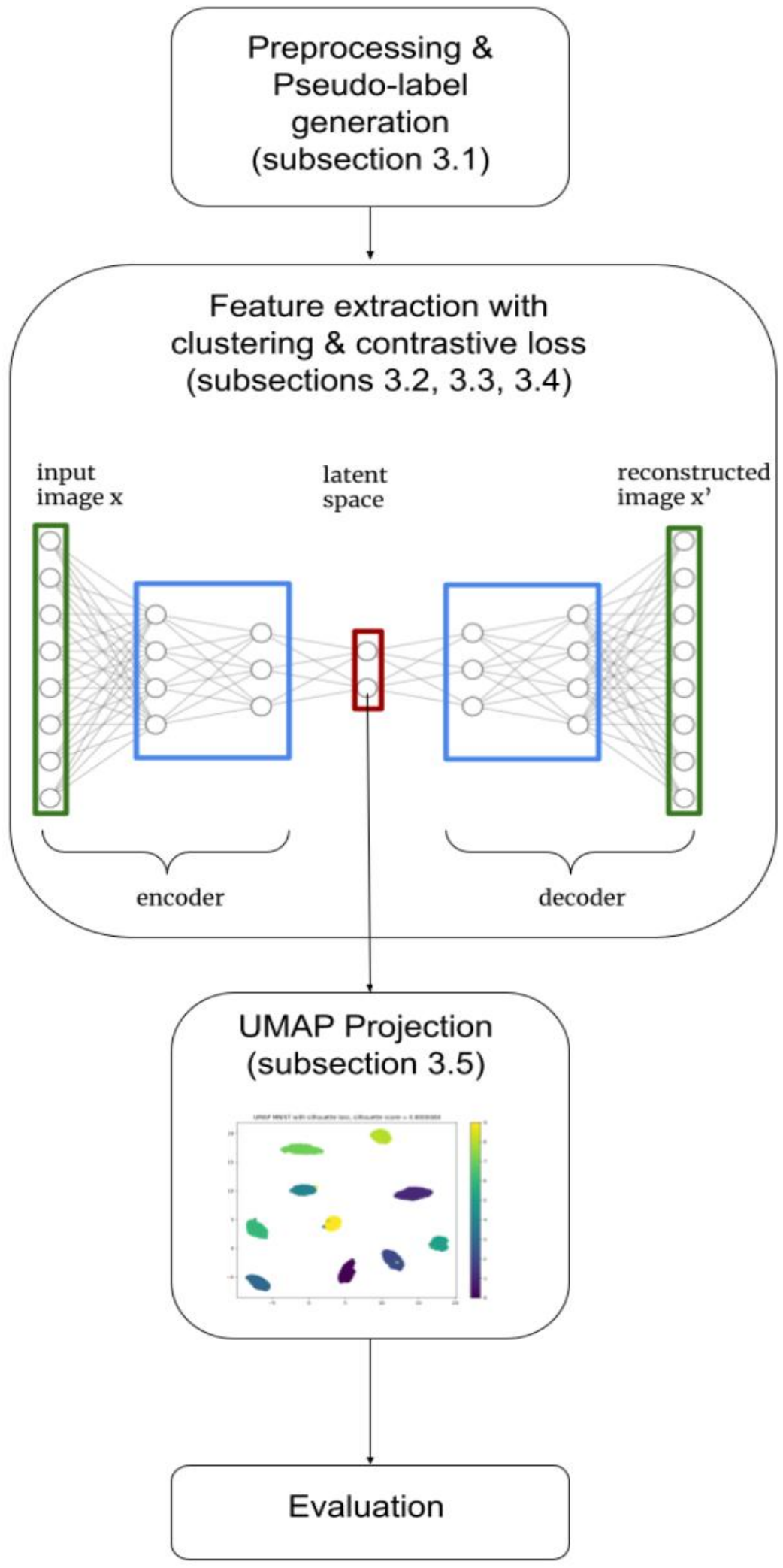

Figure 1: Pipeline of the developed method.

### 3.1 Scientific ensemble datasets

In this project, we consider two ensemble datasets. The first is the Markov chain Monte Carlo (MCMC) ensemble dataset, which depicts channel structures in soil [4]. This ensemble contains 95K images. The images are monochrome and have a resolution of 50 × 50. An example of the MCMC images is shown in Figure 2.

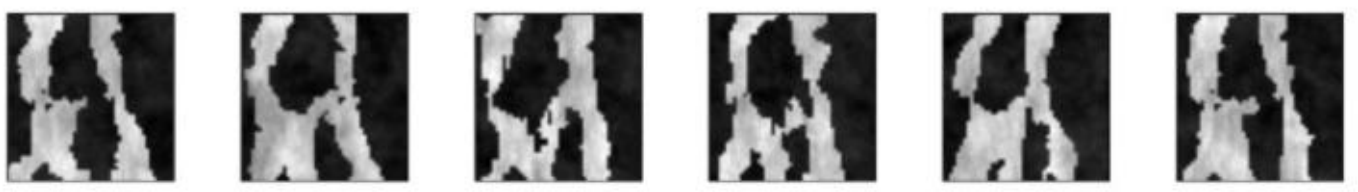

Figure 2: Markov chain Monte Carlo images examples.

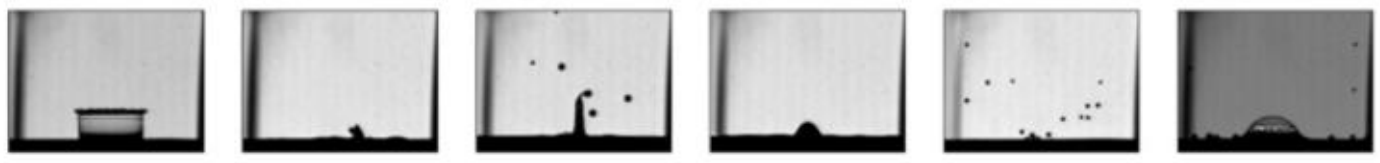

Figure 3: Drop dynamics images examples.

The second ensemble dataset is the drop dynamics (DD) ensemble, which studies the impact of a droplet with a film [5]. These images are also monochrome and have a resolution of 160×224. This ensemble contains 135K images. An example of the DD images is shown in Figure 3.

For both ensembles, a small subset of the datasets is labelled manually. For the MCMC ensemble, 2.5K images are labelled, and 7.2K for the DD ensemble. This labelling was done by observing the different behaviour types shown by the images. For MCMC, there are five different categorical classes. For the DD ensemble, there are seven different categorical classes: bubble, bubble-splash, column, crown, crown-splash, splash, and drop.

Because the ensemble datasets are only partially labelled, we use an EfficientNetV2 model to train on this part, and then generate pseudo-labels for the unlabelled subset of the ensembles, making this a semi-supervised problem. EfficientNetV2 is an image classification convolutional network, and it is known for its small size, speed and performance [20]. For this project, a subset of 30K images for the MCMC ensemble will be used, and 26K images for the DD ensemble. Both datasets are normalized to zero mean and unit standard deviation.

### 3.2 Architecture

In this subsection, (variational) autoencoders will be explained in detail, and the architectures used for this project will be described.

**Autoencoders.** Autoencoders and (β)-variational autoencoders will be used for this project. An autoencoder (AE) is a type of artificial neural network that is often used for unsupervised learning problems [21]. An autoencoder aims to learn a representation (encoding) for a dataset by training the network to reconstruct the original input as accurately as possible. It consists of two main parts: an encoder and a decoder.

The encoder compresses the input data into a lower-dimensional representation, also known as the latent space or encoding. It typically consists of one or more hidden layers that progressively reduce the dimensionality of the input data, mapping it to a lower-dimensional representation. The encoder's output represents a compressed version of the input data, capturing its essential features. Mathematically, an encoder with one hidden layer can be expressed as $z = \sigma(Wx + b)$, where z is the latent representation, σ is the activation function, W is the weight matrix, x is the input image, and b is the bias.

The decoder reconstructs the original input data from the encoded representation produced by the encoder. It typically consists of one or more hidden layers that gradually expand the dimensionality of the encoded representation back to the original input dimensionality. The decoder's output should ideally closely match the input data, effectively "decoding" the compressed representation into a reconstructed version of the original input. Mathematically, a decoder with one hidden layer can be expressed as $x' = \sigma'(W'z + b')$, where $x'$ is the reconstructed image, $\sigma'$ is the activation function, $W'$ is the weight matrix, $z$ is the latent representation, and $b'$ is the bias. The general structure of an autoencoder is shown in Figure 4.

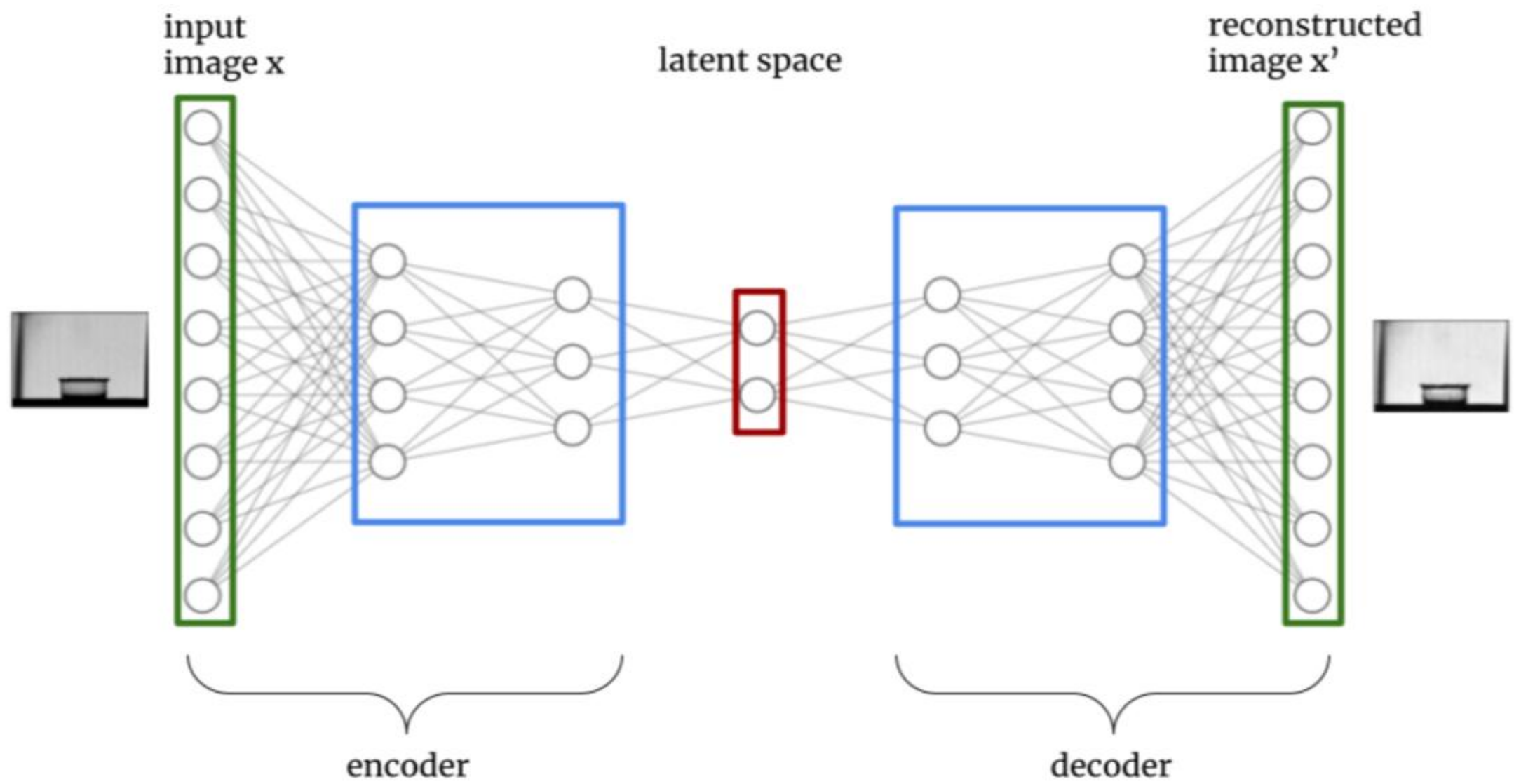

Figure 4: Autoencoder general structure.

During training, an autoencoder is trained to minimize the reconstruction loss function that quantifies the difference between the input data and the reconstructed output. In this project, the Mean Squared Error is chosen as the loss function. Mathematically, MSE is expressed as $L(x, x') = \|x - x'\|_2$.

**Variational autoencoders**. Variational autoencoders (VAE) are a type of autoencoder and are similar to the previously explained autoencoders in many ways [22]. The key difference between a VAE and a traditional autoencoder lies in how they handle the latent space representation. In a traditional autoencoder, the encoder maps input data to a fixed-dimensional latent space, and the decoder reconstructs the input data from this latent representation. However, in a VAE, the latent space is probabilistic, meaning that instead of encoding data points to a specific point z in the latent space, the encoder outputs the parameters of a probability distribution over the latent space. Then, the decoder takes a sampled point from this distribution and generates a new image. Because of VAEs nature of using a sample from the distribution, they can produce new images. This can help in preventing overfitting, as the model learns a more general structure from the image.

Furthermore, VAEs also add a regularization term to the loss function, called the Kullback-Leibler (KL) divergence. The KL divergence encourages the learned latent space to approximate a unit Gaussian distribution, helping to ensure that the latent space remains interpretable. The training objective of a VAE is to maximize the Evidence Lower Bound (ELBO), which is comprised of two parts:

$$\mathbb{E}_{q_\phi(z|x)}[log p_\theta(x|z)] - D_{KL}(q_\phi(z|x)||p_\theta(z)), \quad (1)$$

where the prior distribution with a unit Gaussian distribution (zero mean and unit standard deviation) and the probability distribution of the parameters μ and σ are outputted by the encoder. This means that expected value is equal to the reconstruction loss, and the second part is equal to the KL divergence, which measures the difference between the probability distribution outputted by the encoder and the prior distribution. The ELBO (evidence lower bound) needs to be minimized.

As in the paper "Evaluation and selection of autoencoders for expressive dimensionality reduction of spatial ensembles" [3], the KL divergence is scaled according to the dimensionality of the latent space according to the following formula:

$$\frac{dim(latent)}{height \cdot width}, \tag{2}$$

where dim(latent) is the size of the latent space, and height and width are the height and width of the input image. The general structure of a VAE is shown in Figure 5.

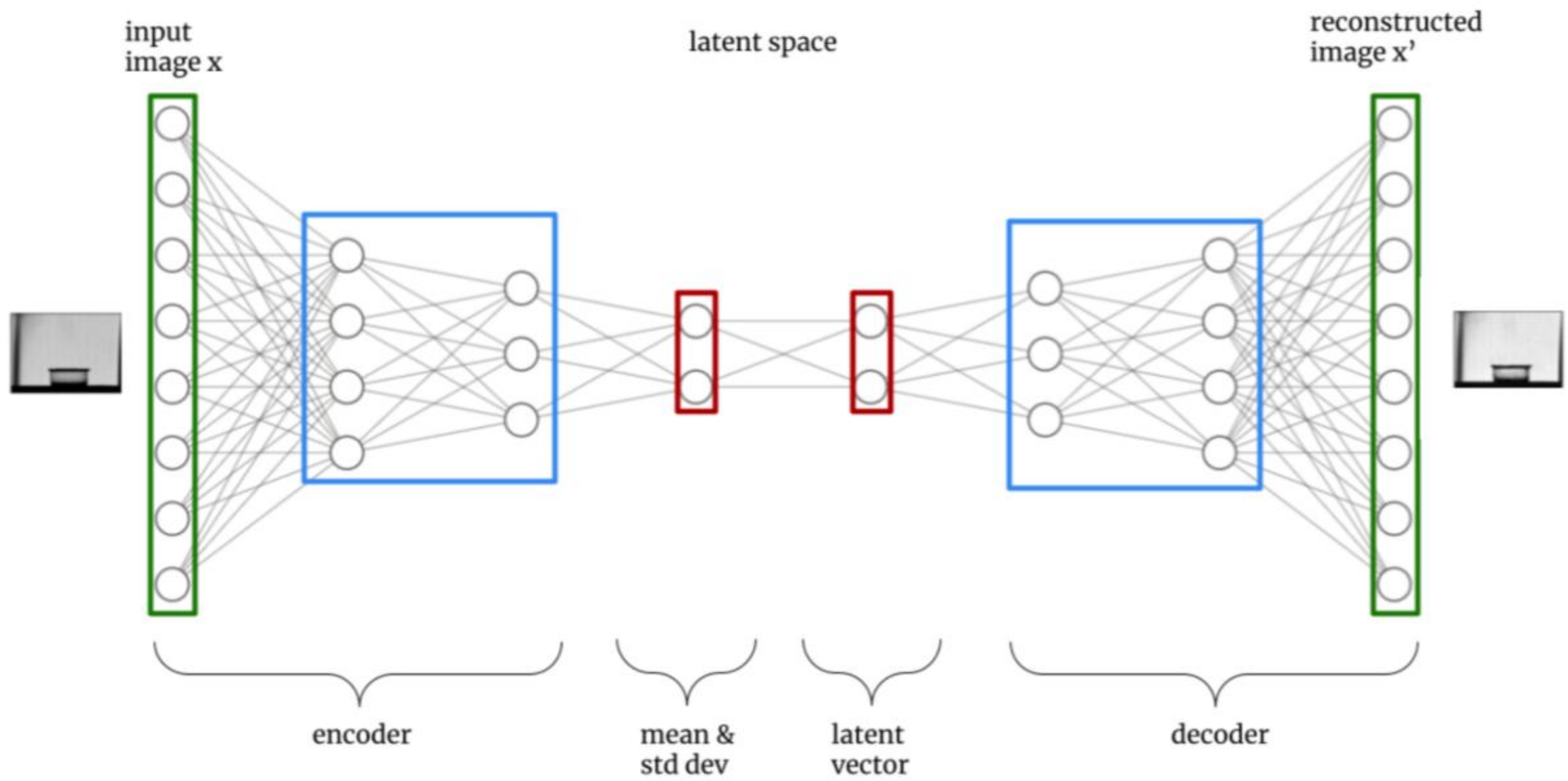

Figure 5: Variational Autoencoder general structure.

A β-VAE is a type of VAE that adds the Lagrangian multiplier β which balances the reconstruction accuracy and impact of the KL divergence factor [23]. Using the Lagrangian multiplier β can help to create an even more expressive latent space. With a β-VAE, the KL divergence coefficient of Equation 2 is multiplied by this Lagrange Multiplier β. This means that a β value of 1 means that it is equal to a VAE.

The architecture used for the AE and (β)-VAE is the same symmetric structure as in [3], meaning that the decoder is reversed to the encoder. First, the resolution of input images is reduced by half after all four convolutional layers. This is done by using a stride of 2. The kernel size for convolution was set to 3, and the number of filters is equal to 64 with zero padding. As an optimizer, Adam was used with a learning rate of 0.0005 [24]. The ReLU activation function was used throughout with random weight initialization. ReLU is linear in the positive dimension but 0 for any negative value [25]. Furthermore, the batch size used in this project is equal to 128. This value was the largest Habrok could reliably handle without running into GPU memory issues.

### 3.3 Clustering loss

To enhance the performance of autoencoder-based clustering, we complement the reconstruction loss with silhouette score as clustering loss. The silhouette score is a widely used clustering quality metric, which measures how similar an item is to other items in the same cluster (cohesion) as well as how different it is from items from other clusters (separation) [26]. The silhouette score ranges from −1 to 1. A high value (close to 1) indicates that an item has good separation from other clusters while also being similar to items from its cluster (good cohesion). A low value (close to -1) indicates that the item has both bad separation and bad cohesion. The silhouette score is defined as follows:

$$s(i) = \frac{b(i) - a(i)}{max(a(i), b(i))}, \tag{3}$$

where a(i) is the mean distance of sample i to all the points in its own cluster, and b(i) is the mean distance between i and the nearest cluster. Thus, to achieve a high silhouette score, a(i) must be minimized, meaning that a point's distance to the other points in its cluster must be low (good cohesion), while b(i) must be maximized, meaning that the distance to other clusters must be high (good separation).

However, we cannot simply implement the silhouette score as is, as the function needs to be differentiable to ensure that backpropagation works. This is why we introduce the soft silhouette score: a differentiable variant of the traditional silhouette score [12]. The soft silhouette score is entirely implemented through tensor operations, ensuring its differentiability. We define the *clustering loss* as follows:

$$\mathcal{L}_{cl} = 1 - \mathcal{S}_f, \tag{4}$$

where Sf is the soft silhouette score. 1 is subtracted from the soft silhouette score to ensure that the clustering loss is in the range [0, 2], where 0 is the best achievable clustering loss. By adding this clustering loss alongside the reconstruction loss Lrec, we ensure that the data will form more compact clusters while also preserving the information of the original data. To strike a balance between the clustering and reconstruction objectives, we introduce a clustering coefficient λcl to scale the contribution of the clustering loss. By adjusting the coefficient of the clustering loss, we control the emphasis placed on clustering relative to reconstruction during training. This approach allows us to fine-tune the trade-off between clustering accuracy and reconstruction performance. This gives us the following loss function:

$$\mathcal{L} = \mathcal{L}_{rec} + \lambda_{cl}\mathcal{L}_{cl} \tag{5}$$

### 3.4 Contrastive loss

Contrastive loss takes the vectors for a positive example and calculates its distance to an example of the same class and contrasts that with the distance to negative examples [27]. The goal is thus to bring instances with the same class closer together while pushing apart the instances with different classes. It helps ensure that the positive examples are represented by similar vectors while negative examples are represented by dissimilar vectors. This is accomplished by calculating the distances of the vectors with any distance metric. For this project, the Euclidean distance formula is used. These distances are then either minimized or maximized, based on whether the samples are positive or negative. The contrastive loss is defined as follows:

$$\mathcal{L}_{con} = y \cdot D^2 + (1-y) \cdot max(0, margin - D)^2, \quad (6)$$

where D is the distance between features, y is 1 if a pair of samples belong to the same class (positive samples), and 0 if not (negative samples). In this formula a margin threshold margin is used. This threshold ensures that negative samples are pushed apart by at least a certain distance. Without the threshold, negative pairs could be arbitrarily close, which would not effectively separate different classes. Moreover, it also helps avoid over-penalizing negative samples that are already far apart. Once the distance between negative samples is larger than the margin, the loss contribution for those pairs becomes zero. In this project, a margin of 1 will be used. Furthermore, to once again strike a balance between the contrastive and reconstruction losses, a contrastive loss coefficient $\lambda$ will also be used to scale the contribution of the contrastive loss. This results in the following loss function:

$$\mathcal{L} = \mathcal{L}_{rec} + \lambda_{con}\mathcal{L}_{con} \quad (7)$$

### 3.5 Projection to 2D Space

Once the ensemble data has been transformed from its original physical space to a feature space through our encoding process, each data sample is represented by a latent vector. These latent vectors are then subjected to dimensionality reduction (DR) techniques. Through DR, the latent vectors are condensed into a lower dimension, in this case, a two-dimensional (2D) representation, while preserving the key information. This transformation to a 2D space helps the visual interpretation of the data, allowing for an informative representation of the patterns present within the dataset.

In this project, we use UMAP (Uniform Manifold Approximation and Projection) to project the latent vectors into a 2D space, as it was shown to outperform other DR techniques such as t-SNE or PCA in the paper "Evaluation and Selection of Autoencoders for Expressive Dimensionality Reduction of Spatial Ensembles" [3]. UMAP does its projection by first determining the similarities between nodes in its original dimension, after which it projects these nodes on a low-dimensional plot [28].

UMAP starts with computing the pairwise distances between the data points in its original dimension. After this, a fuzzy simplicial set is constructed, which captures the local relationship of the data points. Then, the low-dimensional embedding is optimized using stochastic gradient descent. After this optimization process, UMAP provides the 2D representation. Once the projection is done, we evaluate the performance by comparing the silhouette scores. These projections are only done on the manually labelled subset of the ensembles.

### 3.6 Hyperparameter search

In order to get the best results possible, a hyperparameter search is performed to achieve the best-performing model. An initial search is done on the silhouette and contrastive coefficient on the values of {0.01, 0.1, 0.2, 0.3}. The value with the best results is chosen, after which the hyperparameters shown in Table 1 will be searched on the AE and VAE.

In this grid, adaptive weights work as follows: the coefficient for the reconstruction objective will start at 1, whereas the coefficient for the silhouette or contrastive loss will start at 0. With every epoch, the reconstruction coefficient will decrease by 0.01, whereas the other coefficient will increase by 0.01. In case the latent space size is not searched, its default size is 256. The default setup uses no dropout layers.

| Hyperparameter | Values/Used |
|---|---|
| Latent space size | {32, 64, 128, 256} |
| Dropout | {0.2, 0.3, 0.4} |
| $\beta$ (for $\beta$-VAE) | {0.25, 0.5, 0.75, 1, 1.5, 2, 25, 30, 50, 75, 100} |
| Use of learning rate scheduler | {yes, no} |
| Adaptive weights | {yes, no} |
| Train on reconstruction objective, then clustering or contrastive objective | {yes, no} |

Table 1: Grid search hyperparameters.

### 3.7 Setup

All code used for this project was performed using the Python programming language. The (variational) autoencoders were implemented through PyTorch [29]. CUDA was used to train the models on the GPU. The RUG's HPC H´abr´ok was used to perform all experiments on [30]. The amount of epochs is set to 100 for the ensemble datasets training, as we could see both the reconstruction and contrastive or clustering loss converge properly without overfitting. The dataset was split into a training and validation set with an 80%/20% ratio. The training and validation set only use the data that was labelled using the EffNetV2 model. Following training and validation, the UMAP projection is performed on the test set, which is the manually labelled part of the dataset. For the DD ensemble, the categorical classes will be converted into numerical classes to be able to compute the soft silhouette score. After some initial testing, the clustering and contrastive loss coefficients will be set to 0.2, as this value was shown to properly form clusters while succeeding in reconstructing the images.

## 4. Results & Discussion

In this section, we show all the results that we have produced. First, we examine the EffNetV2 pseudo-label generation, then we test our general pipeline on a simple dataset, the MNIST digits dataset. Following that, we show the obtained results from the two ensemble datasets, both with and without contrastive and clustering loss.

**EffNetV2 pseudo-label generation**. Firstly, EffNetV2 was trained on the manually labelled subset of the ensembles. Figure 6 shows that EffNetV2 was properly trained on this subset for both MCMC and DD, achieving a test accuracy of over 95%. Thus, we use this model to generate pseudo-labels for a subset of the unlabelled ensemble datasets.

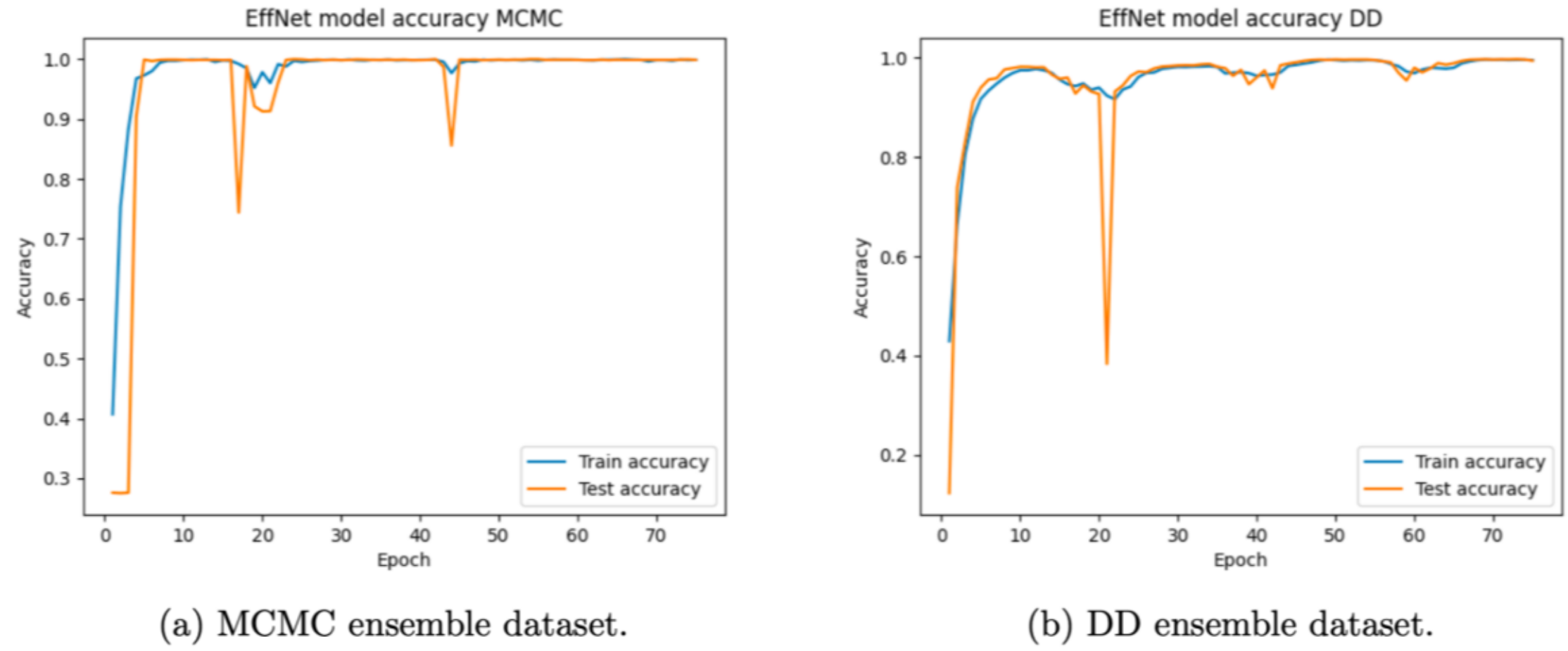

(a) MCMC ensemble dataset.

(b) DD ensemble dataset.

Figure 6: EffNetV2 model accuracy for the ensemble datasets.

**MNIST results.** Following this, the models are constructed and tested on a simple image dataset to see if the task succeeds. We have chosen MNIST: a dataset containing 70000 handwritten digit images of size 28 × 28 [31]. This dataset has 10 classes, representing digits 0 through 9. We compare some results of the AE and the VAE for both with and without silhouette loss and contrastive loss.

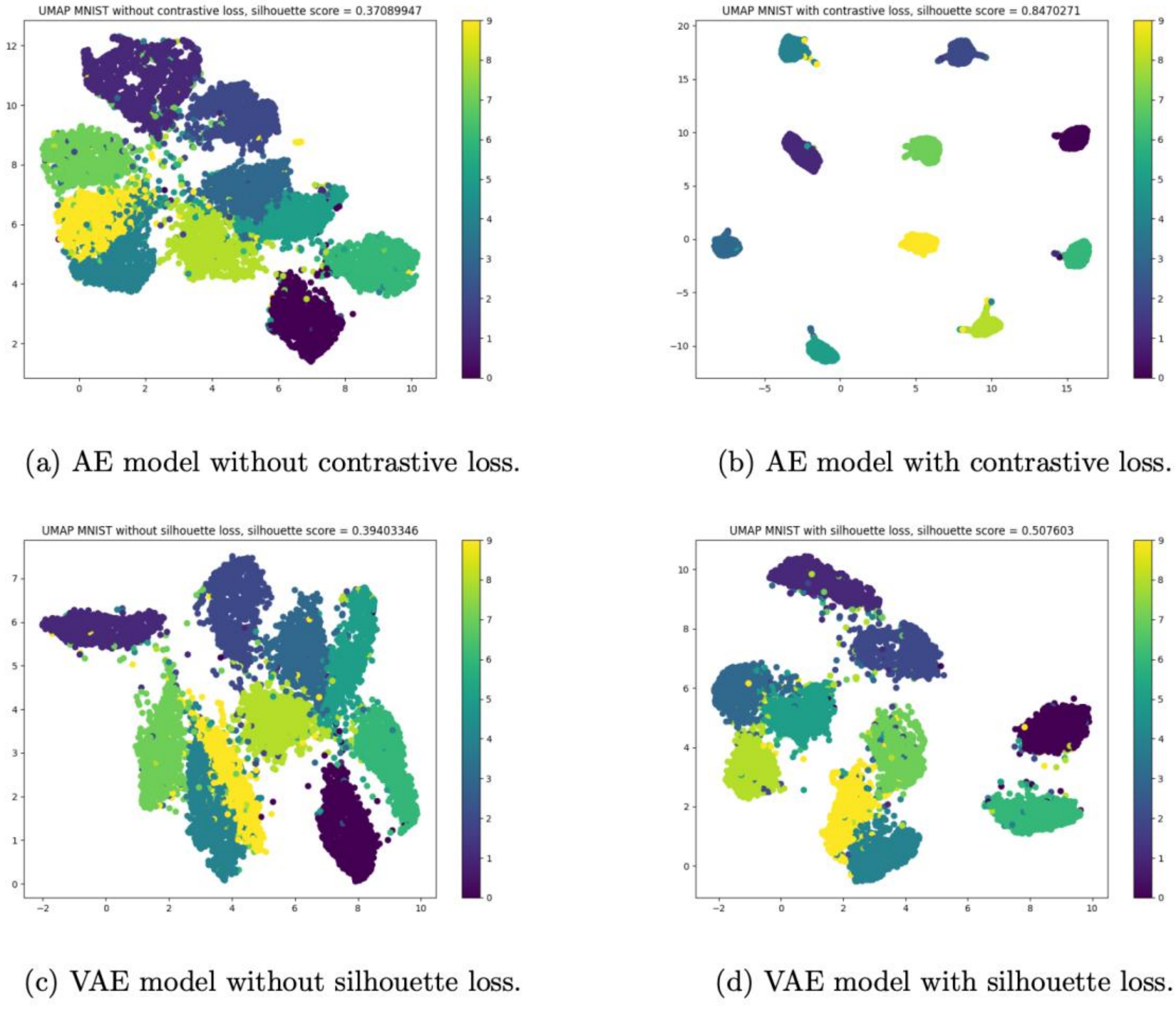

(a) AE model without contrastive loss.

(b) AE model with contrastive loss.

(c) VAE model without silhouette loss.

(d) VAE model with silhouette loss.

Figure 7: MNIST Autoencoder UMAP projections with and without contrastive loss.

Some of the results for this dataset are shown in Figure 7. As we can see, the models with silhouette or contrastive loss outperform the models without these losses and show good separation. In Figure 7b for example, we see that all clusters have some distance from the other clusters. Only a few mistakes are made, such as a few '9's being in the '4' cluster. We notice that the AE model outperforms the VAE model. This is likely due to the VAE's regularization term, as the MNIST dataset is simple enough not to need this. We can see that the AE model is already capable of creating distinct clusters, and thus no more complication of the problem is necessary.

However, we cannot determine which loss or model is superior because no further finetuning has been done yet. We will do this for the scientific ensemble datasets, for which we will perform a hyperparameter search. A note has to be made that the MNIST dataset is a less complicated dataset, for which separation from other clusters is a simple task. This might prove to be more difficult for more complex datasets.

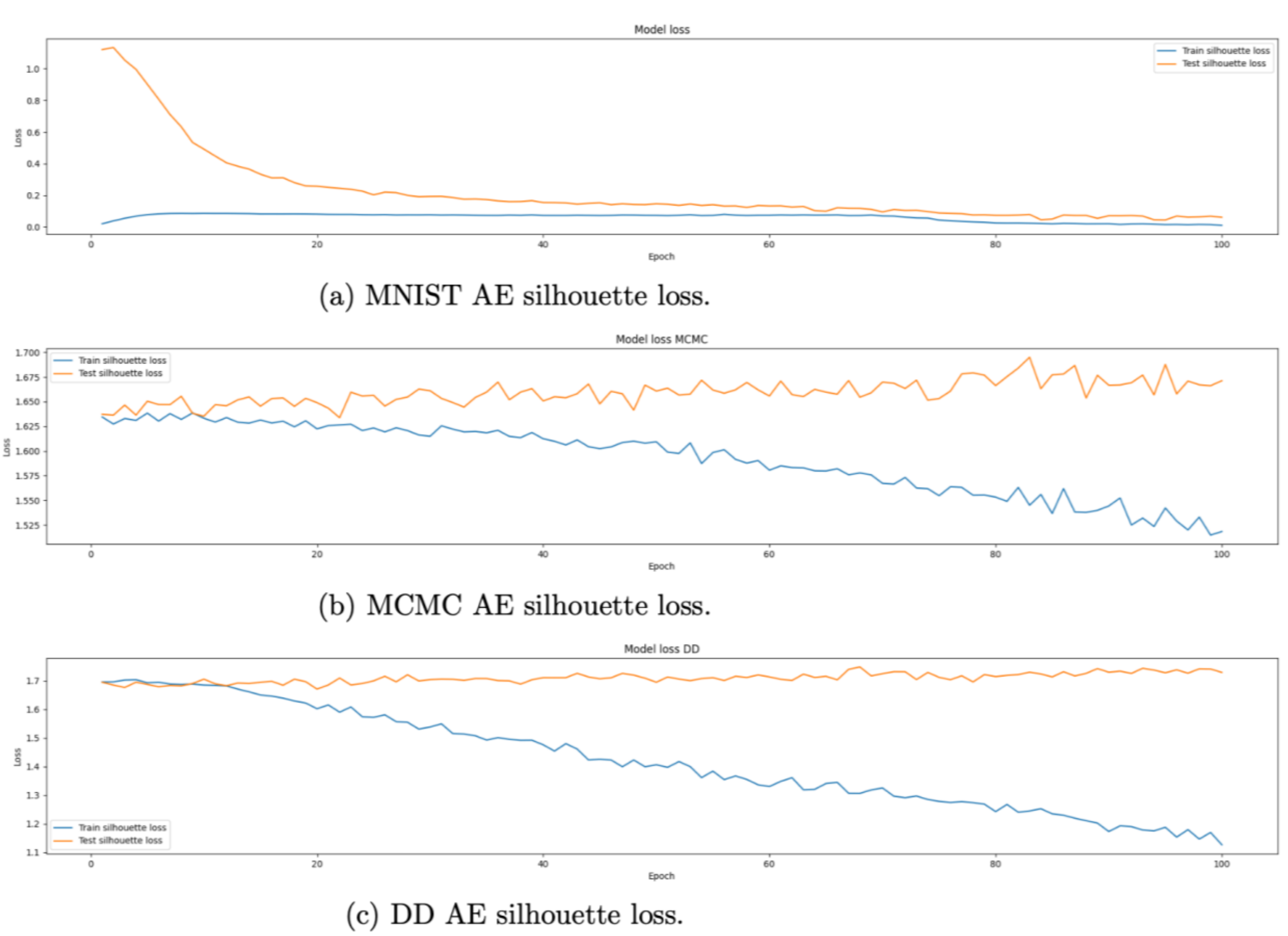

(a) MNIST AE silhouette loss.

(b) MCMC AE silhouette loss.

(c) DD AE silhouette loss.

Figure 8: Autoencoder training and testing silhouette loss.

**Clustering loss issues**. Before heading on to the scientific ensemble datasets in detail, we will first discuss some problems with the clustering loss. During the testing of the entire pipeline for the simpler MNIST dataset, no problems were encountered. Both the soft silhouette loss and the contrastive loss performed as expected. However, with the ensemble datasets, some issues arose. The issue is shown in Figure 8. We notice that for MNIST, shown in (a), the train and test silhouette losses decrease jointly, showing no signs of overfitting. For both MCMC and DD, shown in (b) and (c) respectively, we notice that the train silhouette loss does decrease over time, however, the testing silhouette loss remains high. This is a classic sign of overfitting: the model learns the training data too well and is incapable of generalizing because of it.

A few solutions to overfitting could be to apply early stopping, however, there is no real point at which the models are good at generalization. Batch normalization is also an option, and is tried unsuccessfully. Variational autoencoders apply a regularisation term, which also helps in stopping overfitting. Another solution we tried is to implement dropout: a regularization technique that drops a node in a neural network with a certain probability [32]. However, none of these solutions worked perfectly for the ensemble datasets, which is why we have chosen to primarily show images with contrastive loss, as this all worked adequately. The results with clustering loss are still supplied in the tables that will follow.

An example of what happens to the UMAP projection when the clustering loss is implemented is shown in Figure 9. In this projection, we can see some groups of points forming. However, never is the majority of these points of the same class. So, the model is capable of forming clusters because of the clustering loss, however, because the model is overfitted, it often chooses the wrong samples to cluster.

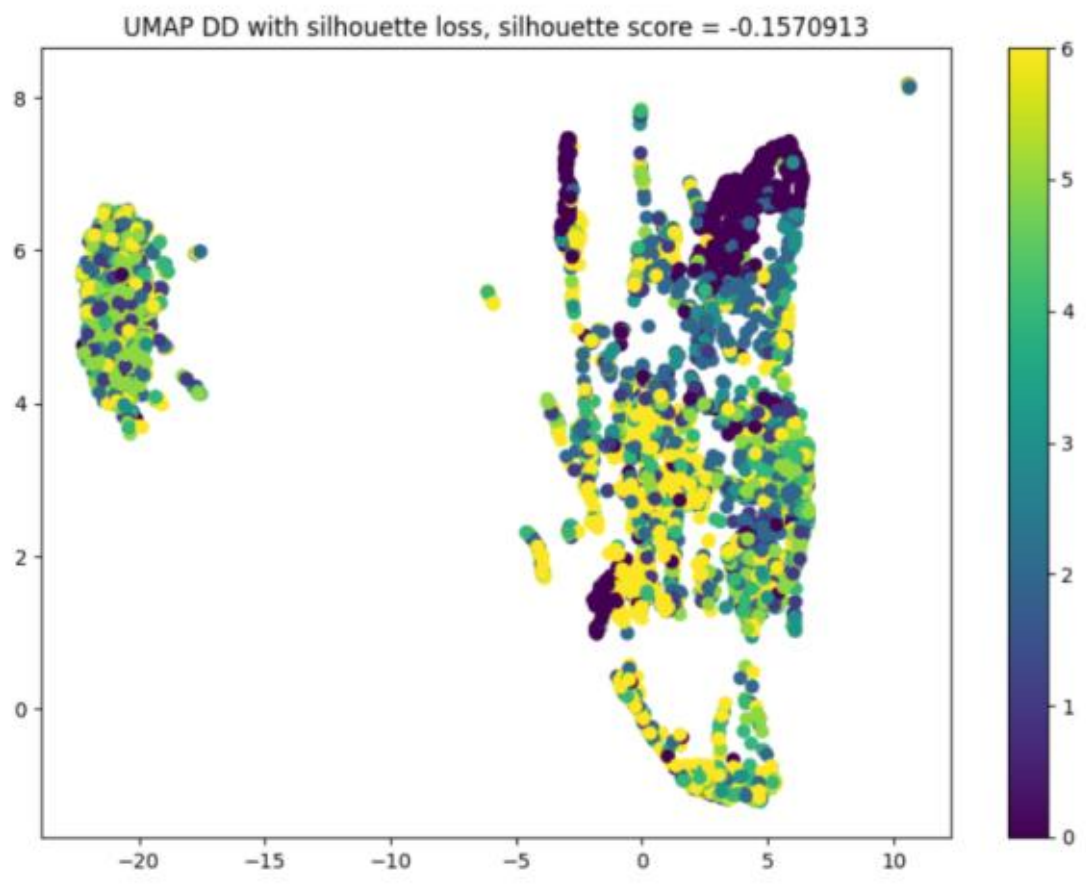

Figure 9: DD Autoencoder UMAP projection with clustering loss.

**Markov chain Monte Carlo ensemble dataset.** Next, we analyze how the models perform on the scientific ensemble datasets. First, we experiment with the Markov chain Monte Carlo ensemble dataset. In Figure 10, we see that both the reconstruction and contrastive objectives converge for the autoencoder models. Figure 10b shows the convergence plots of the model for MCMC, and we notice that the contrastive loss decreases after about 35 epochs and also converges. Compared to Figure 10a the contrastive loss value is significantly higher. We notice that in Figure 10a the test contrastive loss is slightly higher than the training contrastive loss. However, this difference is minimal, so there is no overfitting. Interestingly, the reconstruction losses for both the models with and without con- trastive loss converge in the same way. This shows that the models are capable of performing both the reconstruction and contrastive objectives.

To see if everything works, in Figure 11, the results are shown for a VAE model with and without silhouette loss, with a dropout layer applied with a value of 0.4. We see that the model with the clustering loss does perform better than the model without, albeit marginally. Furthermore, the reconstructed images are very close to the original images. Now, we will observe the results from the hyperparameter search. First, we look at some of the results for MCMC for the autoencoder. These results are shown in Table 2. Please note that in case the latent space size is not searched, its default size is 256. We notice that the best results are produced by the models with either clustering loss or contrastive loss, in this case primarily when dropout of 0.4 is applied. To ensure, we take a look at its reconstructions. These are shown in Figure 12, where we observe that the reconstructions resemble the input images.

| Hyperparameter | Baseline | With clustering loss | With contrastive loss |
|---|---|---|---|
| Latent space=32 | 0.05 | 0.08 | 0.01 |
| Latent space=64 | **0.02** | 0.02 | 0.02 |
| Latent space=256 | 0.03 | 0.05 | −0.03 |
| Dropout=0.3 | 0.03 | 0.07 | −0.01 |
| Dropout=0.4 | 0.03 | **0.10** | **0.14** |
| Pretrained | 0.03 | −0.12 | −0.03 |

Table 2: Autoencoder hyperparameter search: UMAP silhouette scores for MCMC.

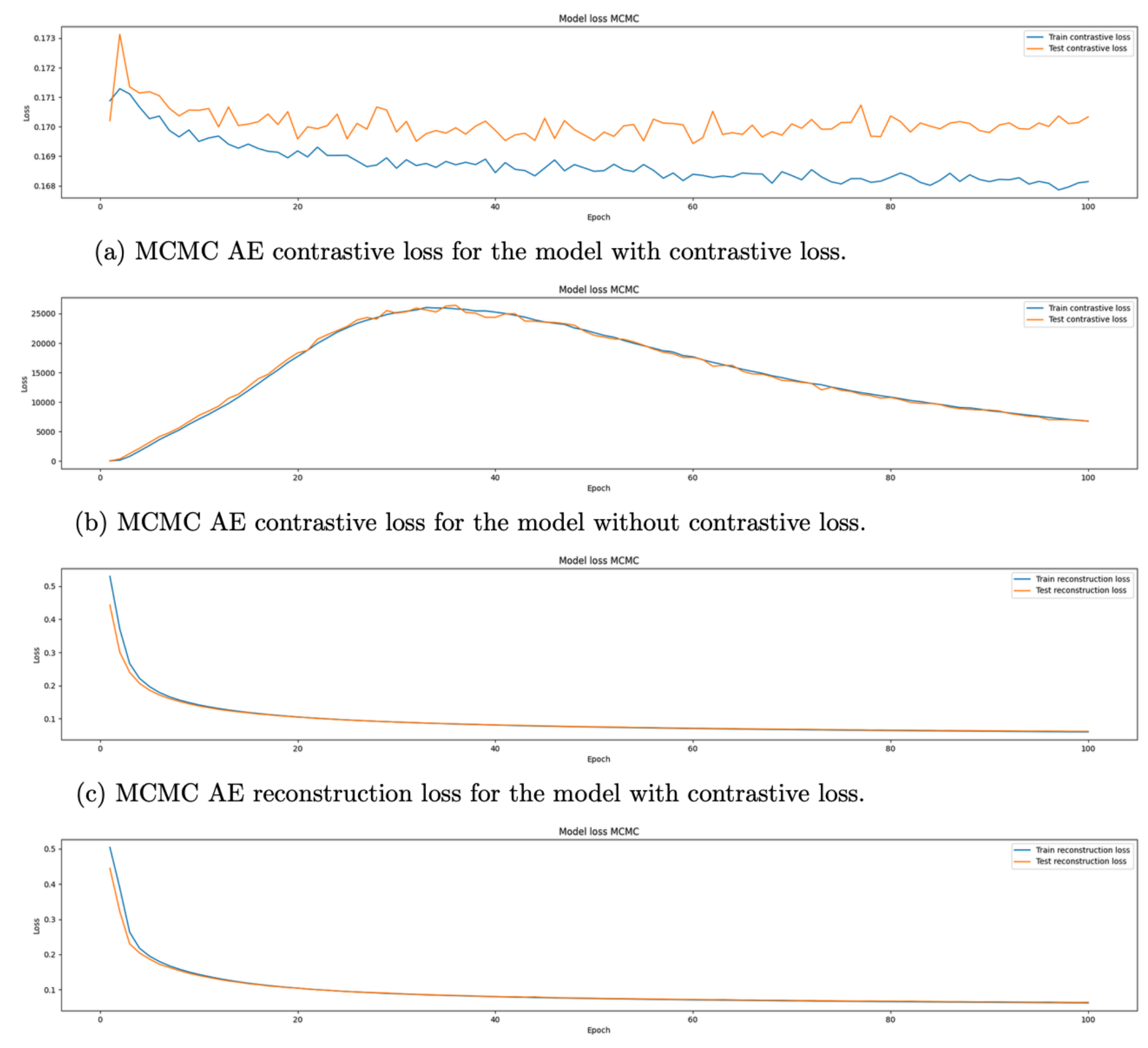

(a) MCMC AE contrastive loss for the model with contrastive loss.

(b) MCMC AE contrastive loss for the model without contrastive loss.

(c) MCMC AE reconstruction loss for the model with contrastive loss.

(d) MCMC AE reconstruction loss for the model without contrastive loss.

Figure 10: MCMC Autoencoder reconstruction and contrastive loss convergence plots for models with and without contrastive loss.

Then, we look at some of the results for MCMC for the (β)-variational autoencoder. These results are shown in Table 3. Here, we immediately see that the results are better than the AE variant, but often only slightly better than the baseline without clustering or contrastive loss. We observe the best result for MCMC, the VAE model with clustering loss with dropout of 0.3.

In Figure 13, a selected few projections are shown that are made for the MCMC ensemble. These projections are all by models with contrastive loss. Firstly, we observe the three autoen- coder model projections, shown in plots (a) through (c). Here, we notice that the difference between the AE model with a latent space of 32 and the model with a latent space of 256 is minimal. The projections show similar clusters and show a minimally different silhouette score. The AE model with a dropout layer with a value of 0.4 shows similar behaviour but shows better cohesion within the clusters.

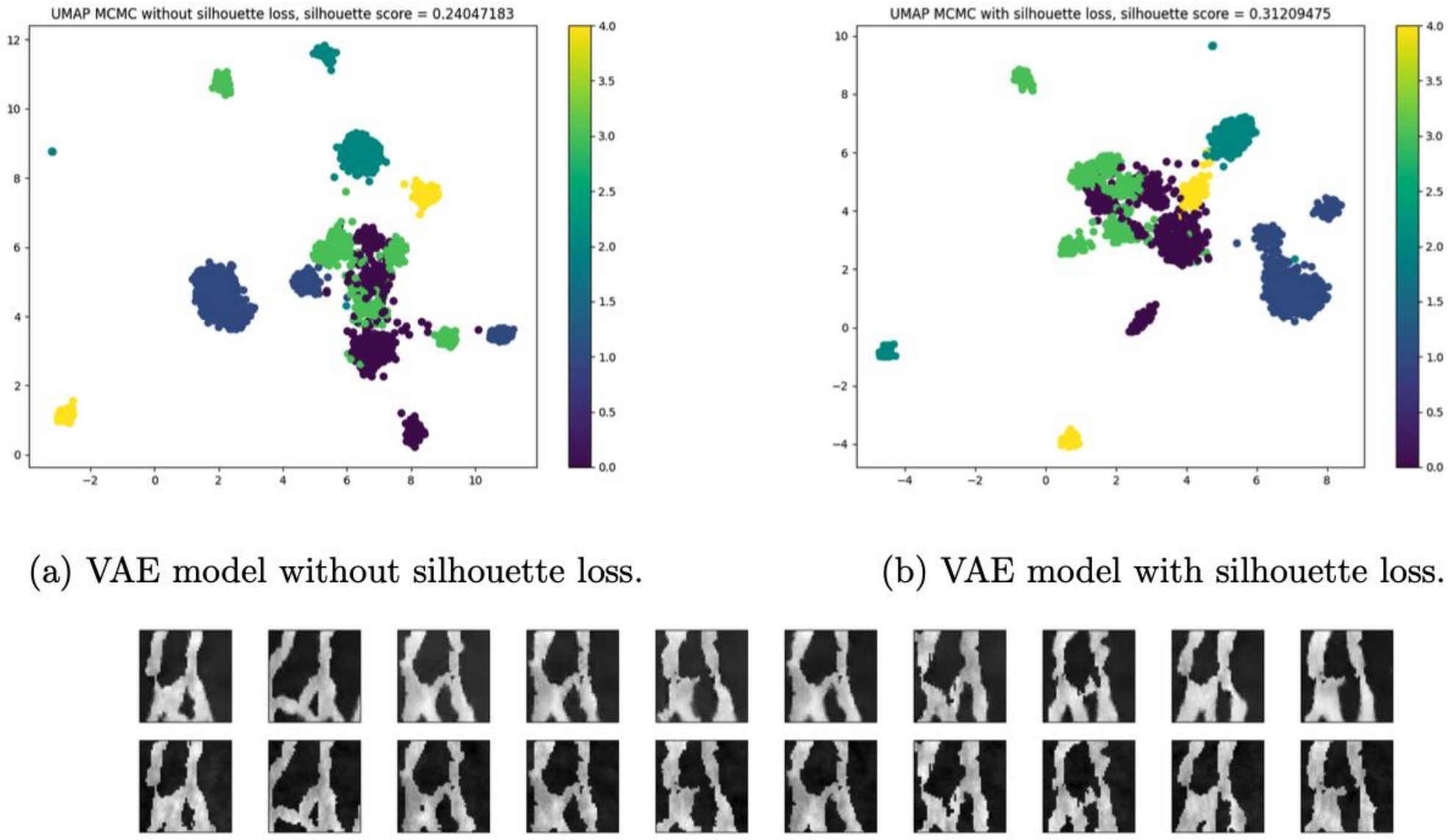

(a) VAE model without silhouette loss.

(b) VAE model with silhouette loss.

(c) VAE model reconstructions (top row) with silhouette loss with original images (bottom row).

Figure 11: MCMC Variational Autoencoder UMAP projections with and without silhouette loss.

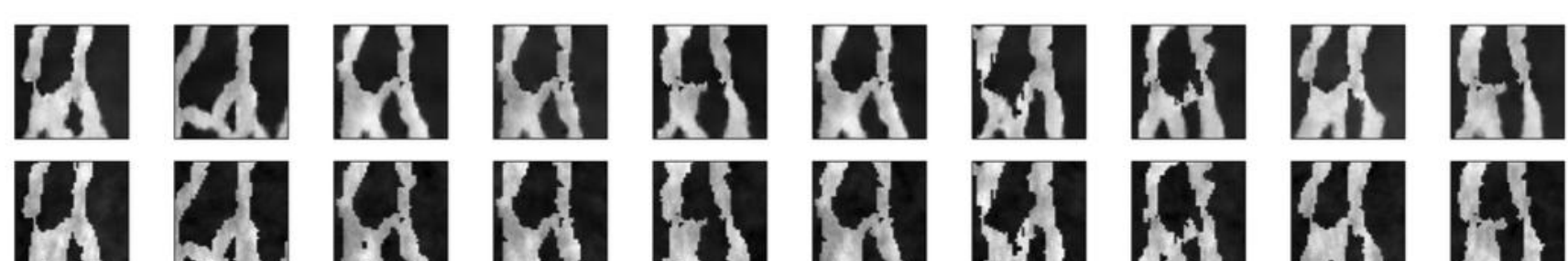

Figure 12: MCMC autoencoder reconstructions (top row) with dropout=0.4 with original images (bottom row).

Then, we analyse the remaining 6 VAE projections. First, we examine plots (d) through (f), the β-VAE models with different Lagrangian multiplier β values. As expected, the β-VAE model with a low value for β, shown in (d), is very similar to the autoencoder projections. This makes sense, as a low value for β decreases the impact of the KL divergence term, causing the clusters to have less of a Gaussian distribution form. However, with a too high value for β, as shown in (e) and ((f)), we notice that all points become scattered together. This means that the KL divergence constraint becomes too important, causing all images to become too general, and thus failing to learn the most relevant features. The latent space should be more constrained.

Now, we look at plots (g) through (i). Comparing (g) with its autoencoder counterpart, shown in (a), we see that the VAE model is more successful in forming cohesive clusters and that the clusters also have more of a Gaussian distribution form. Moreover, although in (g) some of the clusters do not have a lot of separation, some do, such as the cluster in the bottom-right with label 1, as opposed to the top-left cluster with label 2. This effect is not observed as much as in (a).

| Hyperparameter | Baseline | With clustering loss | With contrastive loss |
|---|---|---|---|
| Latent space=32 | −0.01 | 0.00 | 0.05 |
| Latent space=256 | 0.26 | 0.21 | 0.23 |
| Dropout=0.3 | **0.26** | **0.28** | 0.25 |
| Dropout=0.4 | 0.24 | 0.21 | **0.26** |
| $\beta$=0.25 | 0.13 | 0.19 | 0.21 |
| $\beta$=0.5 | 0.06 | 0.13 | 0.19 |
| $\beta$=2 | 0.13 | 0.17 | 0.20 |

Table 3: Variational autoencoder hyperparameter search: UMAP silhouette scores for MCMC.

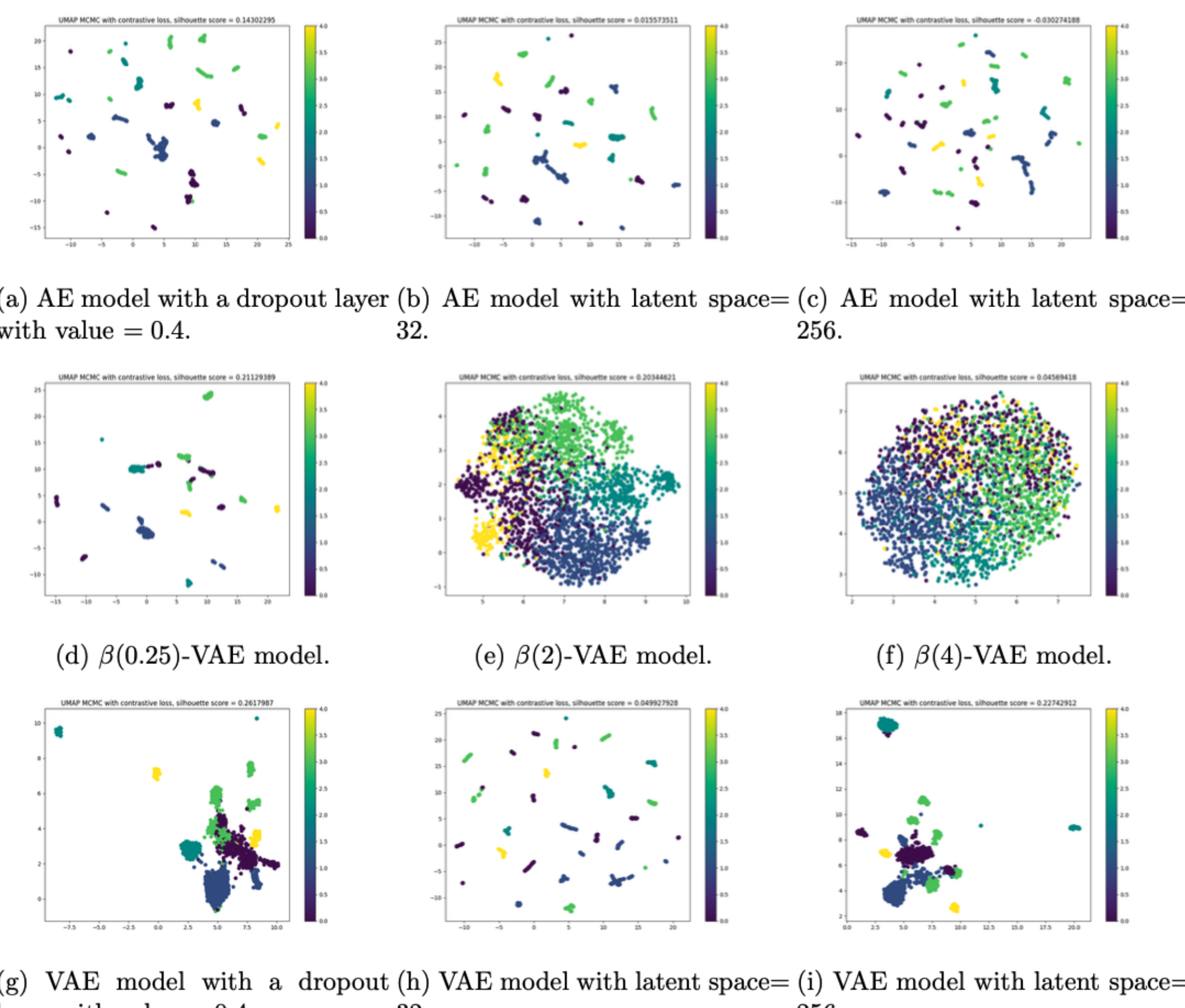

(a) AE model with a dropout layer with value = 0.4.

(b) AE model with latent space= 32.

(c) AE model with latent space= 256.

(d) $\beta$(0.25)-VAE model.

(e) $\beta$(2)-VAE model.

(f) $\beta$(4)-VAE model.

(g) VAE model with a dropout layer with value = 0.4.

(h) VAE model with latent space= 32.

(i) VAE model with latent space= 256.

Figure 13: (($\beta$)-variational) autoencoder models with contrastive loss UMAP projections on MCMC ensemble dataset.

Furthermore, in plots (h) and (i), we compare the effect of the size of the latent space for VAEs. We immediately notice that in (h), the VAE model with a smaller latent space size, the projection is similar to the AE projections. This shows that the bottleneck is too small, causing the VAE struggles to learn relevant features as critical information is lost. The MCMC ensemble dataset is likely too complex to have such a small latent space size. In (i), we see a projection with a higher latent space size of 256. In this projection, we observe the same effect as in (g), with decent cohesion and some separation. Interestingly, comparing (h) and (i) to (b) and (c), we notice that VAEs do benefit from choosing a bigger bottleneck, whereas for AEs, the results are similar no matter the latent space size. This could be because variational autoencoders are more adept at learning complex features due to the KL divergence term.

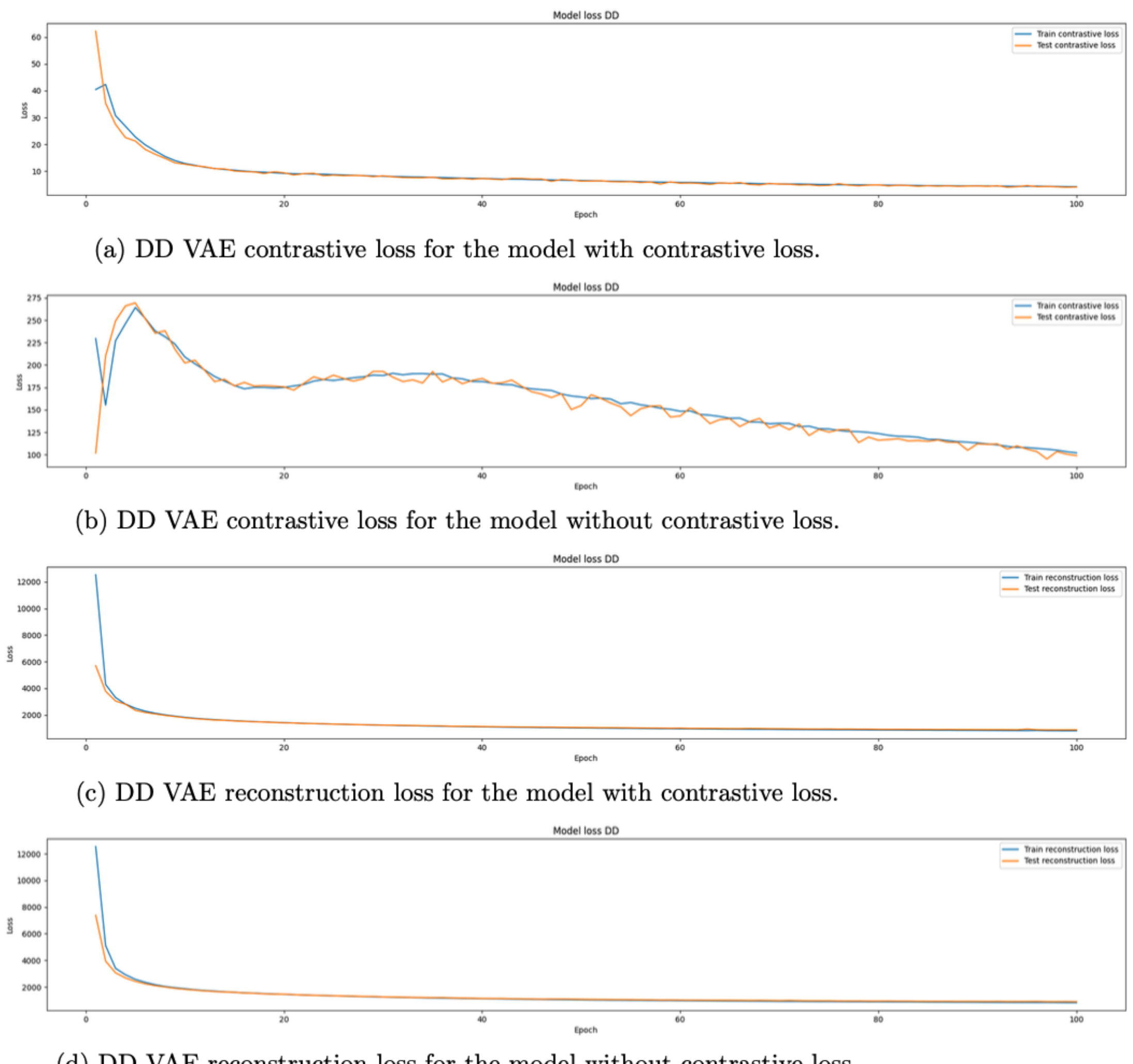

(a) DD VAE contrastive loss for the model with contrastive loss.

(b) DD VAE contrastive loss for the model without contrastive loss.

(c) DD VAE reconstruction loss for the model with contrastive loss.

(d) DD VAE reconstruction loss for the model without contrastive loss.

Figure 14: DD Variational Autoencoder reconstruction and contrastive loss convergence plots for models with and without contrastive loss.

**Drop Dynamics ensemble dataset**. First, in Figure 14a, we show the reconstruction and contrastive losses for Variational Autoen- coder models trained with and without contrastive loss. A VAE is shown here because we already showed the convergence for AE models in the previous subsection, for MCMC. We observe that the reconstruction and contrastive objectives converge for the variational autoencoder models. We notice in (b) that the contrastive losses do decrease over time, however, they are still significantly higher than the contrastive losses in (a). In (c) and (d), we see that the reconstruction losses also converge properly, and are much alike.

Now, we will observe the results from the hyperparameter search for the drop dynamics ensemble dataset. First, we look at some of the results for DD for the autoencoder models. These results are shown in Table 4. We see that the values are quite close to each other, no matter the configurations chosen for the models, and with or without clustering or contrastive loss.

| Hyperparameter | Baseline | With clustering loss | With contrastive loss |
|---|---|---|---|
| Latent space=32 | **-0.08** | −0.09 | −0.11 |
| Latent space=256 | −0.09 | −0.09 | **-0.10** |
| Dropout=0.3 | −0.09 | −0.11 | −0.12 |
| Dropout=0.4 | −0.12 | **-0.07** | −0.14 |
| Pretrained | −0.12 | −0.15 | −0.10 |

Table 4: Autoencoder hyperparameter search: UMAP silhouette scores for DD.

best results here are not significantly better than the autoencoder counterparts.

| Hyperparameter | Baseline | With clustering loss | With contrastive loss |
|---|---|---|---|
| Latent space=32 | −0.09 | −0.13 | −0.11 |
| Latent space=128 | −0.09 | **-0.05** | −0.09 |
| Latent space=256 | −0.08 | −0.11 | −0.10 |
| Dropout=0.3 | −0.08 | −0.12 | −0.12 |
| Dropout=0.4 | −0.08 | −0.08 | −0.11 |
| $\beta$=0.25 | −0.07 | −0.08 | −0.07 |
| $\beta$=1.5 | **-0.06** | −0.11 | −0.08 |
| $\beta$=2 | −0.09 | −0.11 | −0.09 |
| $\beta$=25 | −0.08 | −0.08 | **-0.05** |
| $\beta$=75 | −0.07 | −0.13 | −0.07 |

Table 5: Variational autoencoder hyperparameter search: UMAP silhouette scores for DD.

Now, we will observe the results from the hyperparameter search. First, we look at some of the results for DD for the autoencoder. These results are shown in Table 5. Whereas the results improved for MCMC with the VAE, we cannot say the same for the DD dataset. The best results here are not significantly better than the autoencoder counterparts.

We will analyse the best UMAP projections of both models to see the differences. These projections are shown in Figure 15. We see that the shape of the clusters are quite different, once again most likely that the VAE looks different because of the KL-divergence term causing the clusters to have more of a Gaussian distribution. However, the projection results are not impressive. This is likely due to the extremely complicated nature of the drop dynamics dataset and also due to DD having a large number of classes. The reconstructed images, however, are satisfactory.

In Figure 16, a selected few projections are shown that are made for the DD ensemble. These projections are all by models trained with contrastive loss. In (a) and (b), we compare two autoencoder models. They show that the autoencoder struggles to create clusters, even with the contrastive objective. For the model with the latent space size of 32, we notice that the points are close to being scattered across a 1D line, meaning that the autoencoder possibly simplifies the features too much. This is possible because the number of features in the convolutional layers might be brought down too much, from the images' size of 160 × 224 to a small size of 32 quite quickly, causing too much information to be lost in the process. We see in (b) that this is marginally better for the model with a larger latent space size, however, no real distinct clusters are created either. In (c) through (e), we evaluate the importance of the Lagrangian multiplier β. In (c), we observe more distinct groups of points. However, the complex nature of the dataset makes it difficult to form cohesive clusters which have good separation. This is made even more difficult by having seven different classes. By increasing the value of β, we notice in (e), that the points become fairly general, and that the KL divergence term is too involved. One can notice more of a Gaussian distribution in the created clusters. The clusters' cohesion is quite high, although points of the same class are fairly separate, except for, for example, classes 5 and 6. This behaviour is slightly better in (d), but the UMAP projection's silhouette score is not improved. In (f) and (g), we show the importance of the latent space size for the VAE models. The silhouette score of (g) is a bit better than (f), where the points are separated from each other more. Likely, this is due to the number of features decreasing too much within the model, having to simplify it too much in the process.

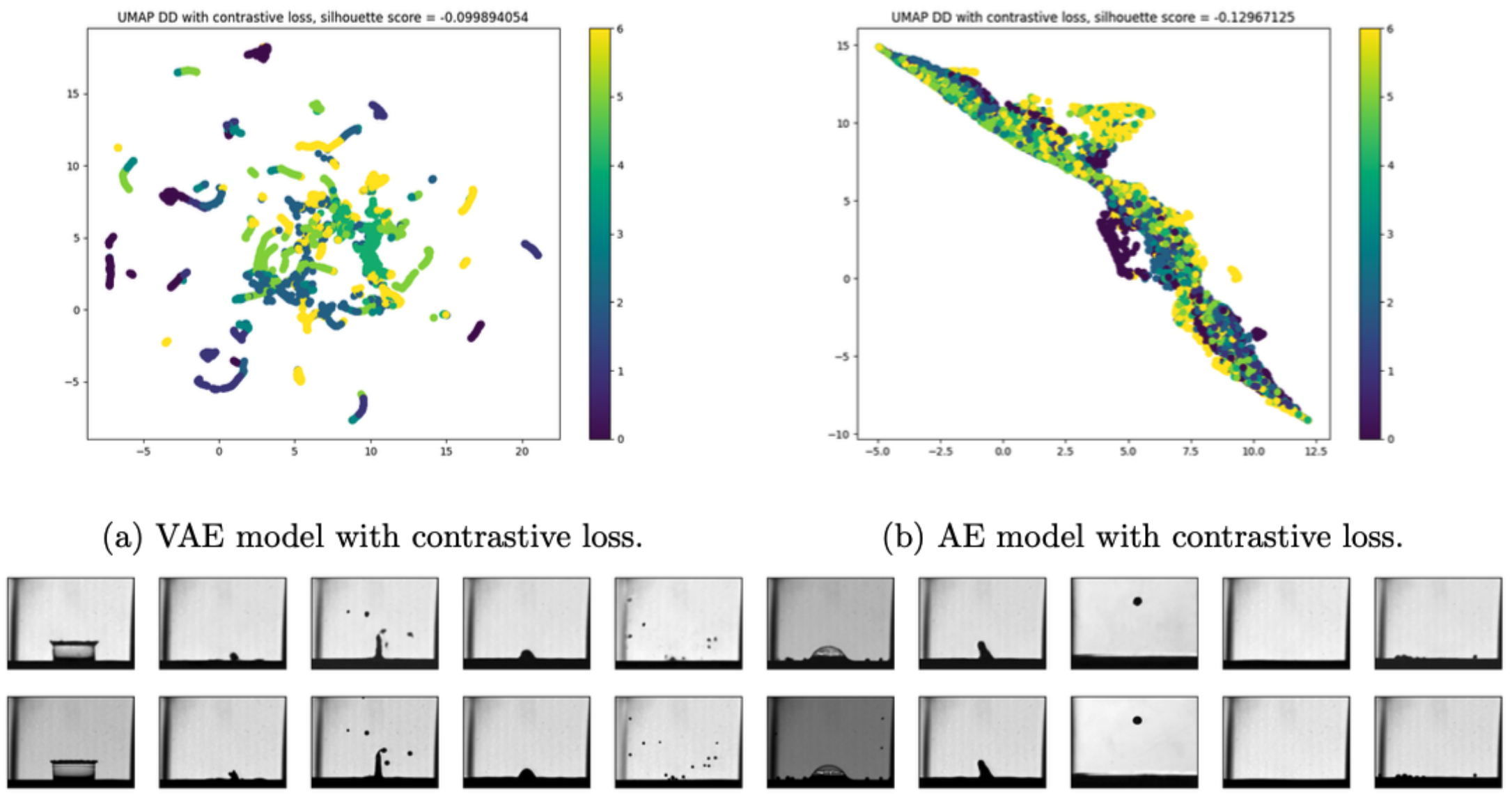

(a) VAE model with contrastive loss.

(b) AE model with contrastive loss.

(c) VAE model with contrastive loss reconstructed images (reconstructions in the top row) with original images (bottom row).

Figure 15: DD Variational- and regular Autoencoder UMAP projections with contrastive loss.

**Comparison of results.** When comparing the results of the MCMC and DD ensemble datasets, we observe that the results for the MCMC dataset are significantly better than for the DD dataset. We suspect this might be because the DD dataset has a few more classes than MCMC does (7 vs. 5) and because the DD dataset contains larger images than MCMC (160 × 224 vs. 50 × 50). Furthermore, DD's differences between classes are often quite minimal, a small splash in the top of the image could be the difference. Also, the crucial parts of these images are often at the bottom of the image, which might be hard to detect for simple distance functions such as the mean squared error or Euclidean distance functions. This is especially different for a simple dataset such as MNIST, where the crucial part of the images are central, and the difference between the classes is easy to observe.

We also observe the difference between the autoencoder models and the (β)-variational autoencoder models. We notice that the (β)-VAE outperform the AE models for both datasets. However, the difference between the two types of models is bigger for the MCMC dataset, where the (β)-VAE models outperform their AE counterparts fairly significantly. This shows the importance of the Kullback-Leibler divergence term. We also notice that the β value has to be significantly higher for the DD dataset to obtain satisfactory results than for the MCMC dataset. This is because of the scaling done in Equation 2. The denominator in this equation is a lot bigger for the DD dataset, as its height and width are larger. This means that a higher β value is necessary to give the KL divergence term the same influence. The KL divergence term helps in creating clusters that have a Gaussian distribution. The importance of the Lagrangian multiplier β is alsoshown multiple times in the figures. As was shown, the models with either clustering or contrastive loss barely outperformed the baselines for both datasets. We suspect, once again, this is due to the datasets' complexity, especially so because it does succeed in the MNIST dataset. For the ensemble datasets, the models succeed in the wanted contrasted effect in the latent space. However, it is too difficult to then project this successfully from a latent space size of 256, for example, to 2D. Unfortunately, too much crucial information is lost in the process.

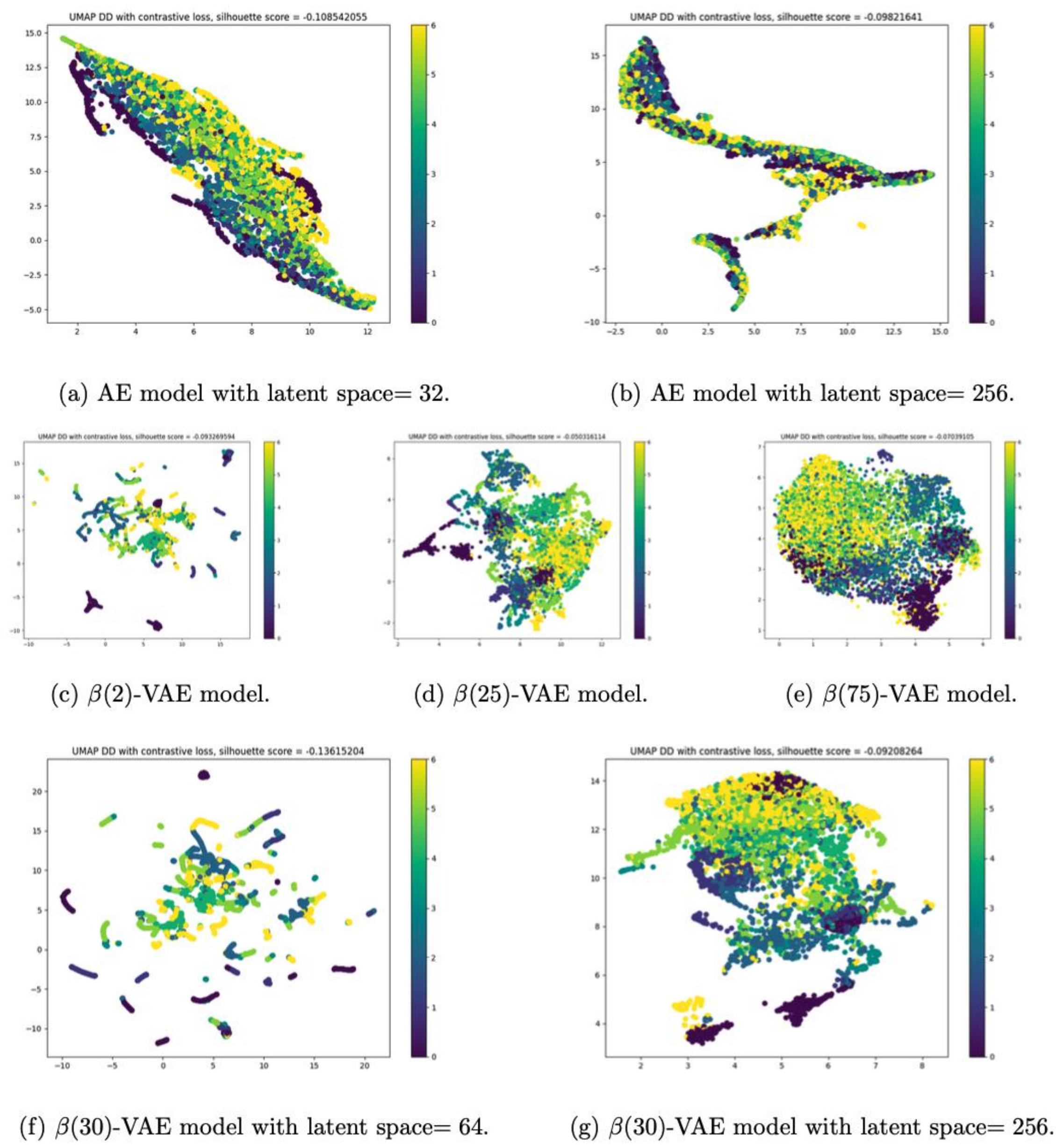

(a) AE model with latent space= 32.

(b) AE model with latent space= 256.

(c) $\beta(2)$-VAE model.

(d) $\beta(25)$-VAE model.

(e) $\beta(75)$-VAE model.

(f) $\beta(30)$-VAE model with latent space= 64.

(g) $\beta(30)$-VAE model with latent space= 256.

Figure 16: (($\beta$)-variational) autoencoder models with contrastive loss UMAP projections on DD ensemble dataset.

## 5. Conclusion

In conclusion, in this project, we have shown different approaches to autoencoder-based semi-supervised dimensionality reduction and clustering for scientific ensembles. We have done this first by generating pseudo-labels for the unlabelled part of the dataset, after which we implemented the soft silhouette score, a clustering loss type, and contrastive loss. Then, with these learning objectives, we compared different configurations of autoencoders and (β)-variational autoencoders, and tested these on two large scientific ensemble datasets, the drop dynamics dataset and the Markov chain Monte Carlo dataset. We projected the latent space to a 2D representation and we obtained results that showed that although it is possible to get better results using clustering or contrastive loss, the improvements are marginal. In general, the β-VAE models outperformed the AE models. For the Markov chain Monte Carlo ensemble, better results were obtained than for the drop dynamics dataset, which is explained by DD's higher number of labels and its respective complexity. Further research has to be done to improve these results.

## 6. Future work

In this section, we will take a look at improvements that could be made to this project, as well as future work that could be implemented. We note some that we believe could be the most promising ones:

- Stopping overfitting with the clustering loss: Clustering loss showed promising results with the simple MNIST dataset. However, it ran into many issues with the scientific ensemble datasets. Some slight changes to the function might make it perform better.
- Combining clustering and contrastive loss with reconstruction loss: Because both loss functions show promising results, it could be interesting to couple them and perform further optimization.
- Improve on drop dynamics results: In this project, it was difficult to significantly improve the results of the drop dynamics ensemble dataset because of its complicated nature. More research should be done to improve upon this.
- Try out other model architectures: Although the models generally outperformed the baselines, there might be some model architecture that outperforms it all. Furthermore, different types of models could be looked into.
- Utilize a different type of loss function: Another way to improve could be to utilize a different type of loss function than contrastive or clustering loss.
- Further analysis of the β value: The Lagrangian multiplier β showed to be quite influential. A larger search for the best value for this could be done. An adaptive weight for this value could also be implemented, changing the value while training.

***Reference list***

1. *Junpeng Wang, Subhashis Hazarika, Cheng Li, and Han-Wei Shen. Visualization and visual analysis of ensemble data: A survey. IEEE Transactions on Visualization and Computer Graphics, 25(9):2853–2872, 2019.*
2. *Julie Jebeile and Michel Crucifix. Multi-model ensembles in climate science: Mathematical structures and expert judgements. Studies in History and Philosophy of Science Part A, 83:44–52, 2020.*
3. *Hamid Gadirov, Gleb Tkachev, Thomas Ertl, and Steffen Frey. Evaluation and selection of autoencoders for expressive dimensionality reduction of spatial ensembles. In Advances in Visual Computing, pp. 222–234, 2021. Springer International Publishing.*
4. *Sebastian Reuschen, Teng Xu, and Wolfgang Nowak. Bayesian inversion of hierarchical geostatistical models using a parallel-tempering sequential gibbs MCMC. Advances in Water Resources, 141:103614, 07 2020.*
5. *Anne Geppert, Dimitrios Chatzianagnostou, Christian Meister, Hassan Gomaa, G. Lamanna, and Bernhard Weigand. Classification of impact morphology and splash- ing/deposition limit for n-hexadecane. Atomization and Sprays, 26, 01 2015.*
6. *Eugen-Richard Ardelean, Andreea Coporˆıie, Ana-Maria Ichim, Mihaela Dˆıns, oreanu, and Raul Cristian Mures, an. A study of autoencoders as a feature extraction technique for spike sorting. PLOS ONE, 18(3):1–29, 03 2023.*
7. *Dillip Ranjan Nayak, Neelamadhab Padhy, Pradeep Kumar Mallick, and Ashish Singh. A deep autoencoder approach for detection of brain tumor images. Computers and Electrical Engineering, 102:108238, 2022.*
8. *Min Chen, Xiaobo Shi, Yin Zhang, Di Wu, and Mohsen Guizani. Deep feature learning for medical image analysis with convolutional autoencoder neural network. IEEE Transactions on Big Data, PP:1–1, 06 2017.*
9. *Enoch Solomon, Abraham Woubie, and Eyael Solomon Emiru. Autoencoder based face verification system, 2024.*
10. *Huiyuan Tian, Li Zhang, Shijian Li, Min Yao, and Gang Pan. Pyramid-vae-gan: Trans- ferring hierarchical latent variables for image inpainting. Computational Visual Media, 9:827–841, 07 2023.*
11. *Xiuxi Wei, Zhihui Zhang, Huajuan Huang, and Yongquan Zhou. An overview on deep clustering. Neurocomputing, 590:127761, 2024.*
12. *Georgios Vardakas, Ioannis Papakostas, and Aristidis Likas. Deep clustering using the soft silhouette score: Towards compact and well-separated clusters, 2024.*
13. *Junyuan Xie, Ross Girshick, and Ali Farhadi. Unsupervised deep embedding for clustering analysis, 2016.*
14. *Xifeng Guo, Long Gao, Xinwang Liu, and Jianping Yin. Improved deep embedded clustering with local structure preservation. 08 2017.*
15. *Bo Yang, Xiao Fu, Nicholas D. Sidiropoulos, and Mingyi Hong. Towards k-means-friendly spaces: Simultaneous deep learning and clustering, 2017.*
16. *Hao Zhou, Ke Yu, Xuan Zhang, Guanlin Wu, and Anis Yazidi. Contrastive autoencoder for anomaly detection in multivariate time series. Information Sciences, 610:266–280, 2022.*
17. *Dawei Luo, Heng Zhou, Joonsoo Bae, and Bom Yun. Combining contrastive learning with auto-encoder for out-of-distribution detection. Applied Sciences, 13:12930, 12 2023.*
18. *Alejo Lopez-Avila and V´ıctor Su´arez-Paniagua. Combining denoising autoencoders with contrastive learning to fine-tune transformer models, 2024.*
19. *Zeyu Cao, Xiaorun Li, Yueming Feng, Shuhan Chen, Chaoqun Xia, and Liaoying Zhao. Contrastnet: Unsupervised feature learning by autoencoder and prototypical contrastive learning for hyperspectral imagery classification. Neurocomputing, 460:71–83, October 2021.*
20. *Mingxing Tan and Quoc V. Le. Efficientnetv2: Smaller models and faster training, 2021.*
21. *Mark A. Kramer. Nonlinear principal component analysis using autoassociative neural networks. Aiche Journal, 37:233–243, 1991.*
22. *Diederik P Kingma and Max Welling. Auto-encoding variational bayes, 2022.*
23. *Irina Higgins, Lo¨ıc Matthey, Arka Pal, Christopher P. Burgess, Xavier Glorot, Matthew M. Botvinick, Shakir Mohamed, and Alexander Lerchner. beta-vae: Learning basic visual concepts with a constrained variational framework. In International Conference on Learning Representations, 2016.*
24. *Diederik P. Kingma and Jimmy Ba. Adam: A method for stochastic optimization, 2017.*
25. *Vinod Nair and Geoffrey Hinton. Rectified linear units improve restricted boltzmann ma- chines vinod nair. volume 27, pages 807–814, 06 2010.*

26. *Peter J. Rousseeuw. Silhouettes: A graphical aid to the interpretation and validation of cluster analysis. Journal of Computational and Applied Mathematics, 20:53–65, 1987.*
27. *R. Hadsell, S. Chopra, and Y. LeCun. Dimensionality reduction by learning an invariant mapping. In 2006 IEEE Computer Society Conference on Computer Vision and Pattern Recognition (CVPR'06), volume 2, pages 1735–1742, 2006.*
28. *Leland McInnes, John Healy, and James Melville. UMAP: Uniform manifold approximation and projection for dimension reduction, 2020.*
29. *Adam Paszke, Sam Gross, Soumith Chintala, Gregory Chanan, Edward Yang, Zachary DeVito, Zeming Lin, Alban Desmaison, Luca Antiga, and Adam Lerer. Automatic differ- entiation in pytorch. In NIPS-W, 2017.*
30. *RUG. High performance computing cluster. https://www.rug.nl/society-business/ centre-for-information-technology/research/services/hpc/facilities/ peregrine-hpc-cluster, 2023.*
31. *Yann LeCun and Corinna Cortes. The MNIST database of handwritten digits. 2005.*
32. *Nitish Srivastava, Geoffrey Hinton, Alex Krizhevsky, Ilya Sutskever, and Ruslan Salakhut- dinov. Dropout: A simple way to prevent neural networks from overfitting. Journal of Machine Learning Research, 15(56):1929–1958, 2014.*
33. *H. Gadirov, Q. Wu, D. Bauer, K. L. Ma, J. B. T. M. Roerdink, & S. Frey, (2025). HyperFLINT: Hypernetwork-based Flow Estimation and Temporal Interpolation for Scientific Ensemble Visualization. Computer Graphics Forum, 44(3), Article e70134. https://doi.org/10.1111/cgf.70134*
34. *H. Gadirov, J. B. T. M. Roerdink and S. Frey, FLINT: Learning-Based Flow Estimation and Temporal Interpolation for Scientific Ensemble Visualization, in IEEE Transactions on Visualization and Computer Graphics, vol. 31, no. 10, pp. 7970-7985, Oct. 2025, doi: 10.1109/TVCG.2025.3561091.*
35. *H. Gadirov. Autoencoder-based feature extraction for ensemble visualization. Master's thesis, University of Stuttgart, 2020. url http://dx.doi.org/10.18419/opus-11304.*
36. *Hamid Gadirov, (2025). Machine Learning for Scientific Visualization: Ensemble Data Analysis. [Thesis fully internal (DIV), University of Groningen]. University of Groningen. https://doi.org/10.33612/diss.1402847307*
37. *Z. Yin, H. Gadirov, J. Kosinka, and S. Frey. ENTIRE: Learning-based Volume Rendering Time Prediction. arXiv preprint arXiv:2501.12119, 2025.*
38. *Bauer, D., Wu, Q., Gadirov, H., & Ma, K. L. (2025). GSCache: Real-Time Radiance Caching for Volume Path Tracing using 3D Gaussian Splatting. IEEE transactions on visualization and computer graphics, PP, 10.1109/TVCG.2025.3634634. Advance online publication. https://doi.org/10.1109/TVCG.2025.3634634*
39. *Hu, E. J., Shen, Y., Wallis, P., Allen-Zhu, Z., Li, Y., Wang, S., ... & Chen, W. (2022). Lora: Low-rank adaptation of large language models. ICLR, 1(2), 3.*
40. *Manuel, L., Gadirov, H., & Frey, S. (2025). Autoencoder-based semi-supervised dimensionality reduction and clustering for scientific ensembles (Research internship project). arXiv. https://arxiv.org/abs/2512.11145*
41. *H. Gadirov. Report on container technology for the ATLAS TDAQ system*
42. *Merkel, D. (2014). Docker: lightweight linux containers for consistent development and deployment. Linux Journal, 2014(239), 2.*